\documentclass[runningheads]{llncs}
\usepackage[T1]{fontenc}

\usepackage{graphicx}
\usepackage{amsfonts}
\usepackage{amsmath}
\usepackage{algorithm}
\usepackage{algpseudocode}
\usepackage{multirow}
\usepackage{color}
\usepackage[colorlinks=true,linkcolor=blue,citecolor=blue,urlcolor=blue]{hyperref}
\begin{document}
\title{Differential privacy representation geometry for medical image analysis}
\author{Soroosh Tayebi Arasteh (1,2)\thanks{Published in MICCAI 2026, Part V, LNCS 16882 proceedings} \and
Marziyeh Mohammadi (3) \and
Sven Nebelung (1,2) \and
Daniel Truhn (1,2)}

\institute{Lab for AI in Medicine, RWTH Aachen University, Aachen, Germany \and 
Department of Diagnostic and Interventional Radiology, University Hospital RWTH Aachen, Aachen, Germany \and
INDA Institute, RWTH Aachen University, Aachen, Germany
\\
\email{* soroosh.arasteh@rwth-aachen.de}}

\maketitle              
\begin{abstract}
Differential privacy (DP)'s effect in medical imaging is typically evaluated only through end-to-end performance, leaving the mechanism of privacy-induced utility loss unclear. We introduce Differential Privacy Representation Geometry for Medical Imaging (DP-RGMI), a framework that interprets DP as a structured transformation of representation space and decomposes performance degradation into encoder geometry and task-head utilization. Geometry is quantified by representation displacement from initialization and spectral effective dimension, while utilization is measured as the gap between linear-probe and end-to-end utility. Across over 594,000 images from four chest X-ray datasets and multiple pretrained initializations, we show that DP is consistently associated with a utilization gap even when linear separability is largely preserved. At the same time, displacement and spectral dimension exhibit non-monotonic, initialization- and dataset-dependent reshaping, indicating that DP alters representation anisotropy rather than uniformly collapsing features. Correlation analysis reveals that the association between end-to-end performance and utilization is robust across datasets but can vary by initialization, while geometric quantities capture additional prior- and dataset-conditioned variation. These findings position DP-RGMI as a reproducible framework for diagnosing privacy-induced failure modes and informing privacy model selection.


\end{abstract}


\section{Introduction}

Deep neural networks in medical image analysis are trained on highly sensitive patient data \cite{mohammadi2026differential}. Although such models achieve state-of-the-art diagnostic performance, they may memorize individual-specific patterns, raising concerns about membership inference, reconstruction attacks, and regulatory compliance \cite{66kaissis2021end,22tayebi2024preserving,usynin2021adversarial}. Differential privacy (DP)~\cite{dwork2014algorithmic} provides a formal guarantee that limits the influence of any single patient on the learned model. A randomized algorithm $\mathcal{A}$ satisfies $(\varepsilon,\delta)$-DP if for all neighboring datasets $\mathcal{D},\mathcal{D}'$ differing in one sample and all measurable outputs $\mathcal{S}$,

\begin{equation}
\Pr[\mathcal{A}(\mathcal{D})\in\mathcal{S}]
\leq
e^{\varepsilon}
\Pr[\mathcal{A}(\mathcal{D}')\in\mathcal{S}]
+
\delta.
\end{equation}
Smaller $\varepsilon$ implies stronger privacy. In deep learning, DP is typically implemented \cite{66kaissis2021end,22tayebi2024preserving,67ziller2024reconciling,mohammadi2026differential} via DP-SGD~\cite{DPSGD}, which clips per-sample gradients and injects Gaussian noise. While this ensures provable privacy, it perturbs optimization dynamics and often reduces predictive performance.

In medical imaging, this privacy-utility trade-off is almost exclusively evaluated through end-to-end task metrics such as AUROC or Dice \cite{66kaissis2021end,mohammadi2026differential,22tayebi2024preserving,13TayebiDomainTransfer,72arasteh2024differentialprivacyenablesfair}. Yet models are rarely used once; they are fine-tuned, transferred across institutions, or deployed as frozen feature extractors. End-to-end performance alone does not reveal whether privacy noise reduces linear separability, reshapes representation geometry, or primarily impairs optimization of the task head \cite{10555535249383525087}, so privacy model selection remains empirical rather than diagnostic.
Representation geometry provides a principled perspective on this problem. The covariance spectrum of embeddings characterizes intrinsic dimensionality and anisotropy, and privacy-constrained optimization can induce structured spectral reshaping rather than uniform collapse \cite{NEURIPS2019_cfcce062}. What is missing is a framework connecting such geometric changes to downstream utility under DP.

We address this gap by introducing the \textbf{D}ifferential \textbf{P}rivacy \textbf{R}epresentation \textbf{G}eometry for \textbf{M}edical \textbf{I}maging (DP-RGMI) framework. DP-RGMI interprets DP training as a transformation of representation space and separates geometric change of the encoder from utilization by task head. Concretely, it quantifies (i) representation displacement from a shared pretrained initialization, (ii) spectral effective dimension of the learned embeddings, and (iii) a utilization gap defined as the difference between linear-probe AUROC and end-to-end private AUROC.

\begin{figure}[t]
\centering
\includegraphics[width=0.91\textwidth]{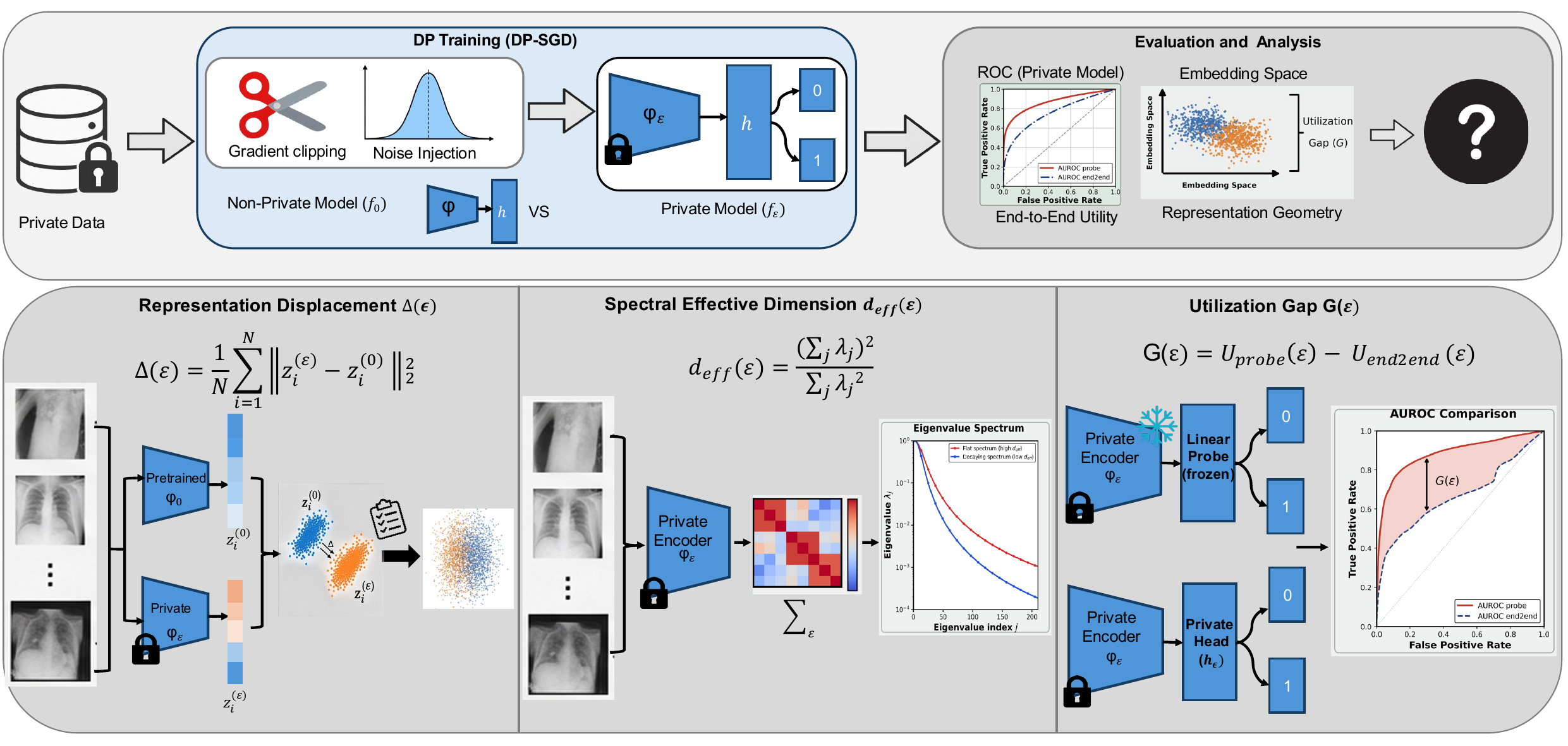}
\caption{Overview of DP-RGMI framework decomposing DP training into representation displacement $\Delta(\varepsilon)$, spectral structure $d_{\mathrm{eff}}(\varepsilon)$, and utilization gap $G(\varepsilon)$.}
\label{fig:framwork}
\end{figure}


\section{DP-RGMI framework}
\label{sec2}

We formalize DP as a transformation of representation space rather than only a scalar constraint on predictive performance. Given a pretrained encoder $\phi_0$ and its differentially private counterpart $\phi_\varepsilon$, DP-RGMI decomposes the analysis into three components (Fig.~\ref{fig:framwork}): representation displacement, spectral structure, and utilization, separating what remains encoded in $\phi_\varepsilon$ from how effectively it is exploited during private training.
We consider a model factorized as $f_\varepsilon(x)=h_\varepsilon(\phi_\varepsilon(x))$, where $\phi_\varepsilon:\mathcal{X}\rightarrow\mathbb{R}^d$ is the encoder and $h_\varepsilon$ a linear head. All geometric quantities are defined in the embedding space of $\phi_\varepsilon$ and compared to the fixed initialization $\phi_0$, giving a shared coordinate system across privacy regimes. Algorithm~\ref{alg:dprgmi} summarizes the workflow.

\begin{algorithm}[t]
\caption{DP-RGMI workflow}
\label{alg:dprgmi}
\begin{algorithmic}[1]
\Require Probe regularization $\lambda$; utility metric $U(\cdot)$
\Ensure For each $\varepsilon$: $(U_{\text{end2end}},U_{\text{probe}},G,\Delta,d_{\mathrm{eff}})$

\For{$\varepsilon \in \mathcal{E}\cup\{\infty\}$}
    \State Train $(\phi_\varepsilon,h_\varepsilon)$ using DP-SGD with budget $\varepsilon$
    \State $U_{\text{end2end}}(\varepsilon)\gets U(h_\varepsilon\circ\phi_\varepsilon;\mathcal{D}_{\text{test}})$
    \State $Z_\varepsilon\gets [\phi_\varepsilon(x_i)]$, $Z_0\gets [\phi_0(x_i)]$ on $\mathcal{D}_{\text{test}}$
    \State $\Delta(\varepsilon)\gets \frac{1}{N}\sum_{i=1}^{N}\|Z_\varepsilon[i]-Z_0[i]\|_2^2$
    \State $\Sigma_\varepsilon\gets \frac{1}{N}(Z_\varepsilon-\mu_\varepsilon)^\top(Z_\varepsilon-\mu_\varepsilon)$
    \State $d_{\mathrm{eff}}(\varepsilon)\gets \frac{\mathrm{tr}(\Sigma_\varepsilon)^2}{\mathrm{tr}(\Sigma_\varepsilon^2)}$
    \State Train linear probe $\hat{h}_\varepsilon$ on frozen $\phi_\varepsilon$
    \State $U_{\text{probe}}(\varepsilon)\gets U(\hat{h}_\varepsilon\circ\phi_\varepsilon;\mathcal{D}_{\text{test}})$
    \State $G(\varepsilon)\gets U_{\text{probe}}(\varepsilon)-U_{\text{end2end}}(\varepsilon)$
\EndFor
\State \Return geometric diagnostic profile of DP training
\end{algorithmic}
\end{algorithm}

\subsubsection{Representation displacement.}

Let $z_i^{(\varepsilon)}=\phi_\varepsilon(x_i)$ and $z_i^{(0)}=\phi_0(x_i)$ denote embeddings of the same test samples under private and initial encoders. 
We quantify representation displacement as:
\begin{equation}
\Delta(\varepsilon)
=
\frac{1}{N}
\sum_{i=1}^{N}
\|z_i^{(\varepsilon)}-z_i^{(0)}\|_2^2.
\end{equation}
This measures how strongly DP-constrained optimization deviates from pretrained prior. 
Crucially, $\Delta(\varepsilon)$ captures geometric movement independently of task labels and isolates privacy-induced change from task-specific fitting. We use a Euclidean metric to keep $\Delta$ orthogonal to the spectral diagnostic: a covariance-aware metric (e.g., Mahalanobis) would whiten by the same covariance whose anisotropy $d_{\mathrm{eff}}$ characterizes, conflating the two.

\subsubsection{Spectral structure.}

Let $\Sigma_\varepsilon=\frac{1}{N}\sum_{i=1}^{N}(z_i^{(\varepsilon)}-\mu_\varepsilon)(z_i^{(\varepsilon)}-\mu_\varepsilon)^\top$ denote embedding covariance with eigenvalues $\{\lambda_j\}$. 
We compute the effective dimension as:
\begin{equation}
d_{\mathrm{eff}}(\varepsilon)
=
\frac{\left(\sum_j \lambda_j\right)^2}{\sum_j \lambda_j^2}.
\end{equation}
This quantity summarizes spectral concentration and anisotropy. 
Changes in $d_{\mathrm{eff}}$ reflect how DP reshapes variance distribution across principal directions rather than merely translating embeddings \cite{NEURIPS2019_cfcce062}.

\subsubsection{Utilization.}

To decouple intrinsic separability from private joint optimization, we freeze $\phi_\varepsilon$ and train a regularized linear probe. Probe utility $U_{\mathrm{probe}}$ measures linear recoverability of class structure in the embedding. The utilization gap is defined as:
\begin{equation}
G(\varepsilon)
=
U_{\mathrm{probe}}(\varepsilon)
-
U_{\mathrm{end2end}}(\varepsilon),
\end{equation}
which quantifies performance loss attributable to optimization under DP rather than representational collapse. In this study $U=\mathrm{AUROC}$, but the definition is metric-agnostic.
A large $G(\varepsilon)$ indicates that discriminative structure persists in $\phi_\varepsilon$ but is not fully exploited during private training.

DP-RGMI is model-agnostic and dataset-agnostic: it requires only access to embeddings and standard evaluation metrics.


\section{Experimental setup}

\subsubsection{Data.}

We study multi-label chest X-ray (CXR) classification on PadChest \cite{bustos2020padchest} (110,525 frontal images from 67,205 patients) as the primary dataset, and use an additional 269,796 images from other public CXR datasets for generalization. PadChest was chosen for primary analysis as it provides binary presence/absence annotations, the most radiologist-annotated labels among available datasets, and sufficient size for stable geometric evaluation.
As no official split exists, we construct a fixed \textit{patient-stratified} partition into training, validation, and test sets. All geometric and utility analyses are performed exclusively on the held-out test set (22,045 images). We focus on five common findings: atelectasis, cardiomegaly, pleural effusion, pneumonia, and no finding. Images are resized to $224\times224$, intensity-normalized, and contrast-standardized following prior work \cite{13TayebiDomainTransfer,arasteh2026roleselfsupervisedpretrainingdifferentially}. Class imbalance is handled via label-wise loss weighting. Code and pretrained weights are publicly available \footnote{\url{https://github.com/tayebiarasteh/CXR-adaptation}, and weights are available from HuggingFace.}.

\subsubsection{Model and training.}

We use ConvNeXt-Small \cite{Liu_2022_CVPR} ($49$M parameters, embedding dimension $d=768$) with a linear multi-label head. ConvNeXt avoids batch normalization, which is incompatible with the per-sample gradients required by DP-SGD, and remains stable under gradient clipping and additive noise; convolutional backbones have predominated in DP-SGD medical imaging, whereas transformers \cite{vaswani2017attention} show unstable DP optimization \cite{mohammadi2026differential}.
Models are optimized with AdamW (weight decay $0.01$). Non-private training uses learning rate $10^{-5}$, batch size $128$, weighted binary cross-entropy loss, and light augmentation (random horizontal flips and small rotations).

\subsubsection{DP training.}

Private runs use DP-SGD without data augmentation \cite{mohammadi2026differential,22tayebi2024preserving}. Per-sample gradients $g_i=\nabla_\theta\ell(\theta;x_i,y_i)$ are clipped to $\ell_2$ norm $C$ via $\bar{g}_i=g_i\cdot\min\!\left(1,\frac{C}{\|g_i\|_2}\right)$ and perturbed with Gaussian noise,
$\tilde{g}=\frac{1}{|\mathcal{B}|}\sum_{i\in\mathcal{B}}\bar{g}_i+\mathcal{N}(0,\sigma^2 C^2 I)$,
followed by the update $\theta\leftarrow\theta-\eta\tilde{g}$, with clipping norm $C=4$.
Training uses Poisson subsampling: each example is independently included in a batch with probability $q=128/|\mathcal{D}_{\text{train}}|$, consistent with privacy accounting. A Privacy Loss Random Variable (PRV) DP accountant \cite{NEURIPS2021_6097d8f3} tracks $(\varepsilon,\delta)$ with $\delta=6\times10^{-6}$ fixed, which satisfies $\delta<1/|\mathcal{D}_{\text{train}}|$ for all datasets; $\varepsilon$ is controlled by adjusting the noise multiplier $\sigma$. Each initialization branch includes a non-private baseline ($\varepsilon=\infty$) and 3 decreasing privacy budgets, all within $\varepsilon<10$, a commonly adopted privacy range in medical imaging studies as a private model \cite{mohammadi2026differential}. All private runs use a maximum budget of $150$ epochs; since the accountant accumulates privacy expenditure per step, each run is trained until convergence and the reported $\varepsilon$ corresponds to its selected checkpoint, which differs across runs.
Privacy guarantees are applied at the image level; since training samples correspond to individual radiographs, the formal guarantee applies per image rather than per patient \cite{13TayebiDomainTransfer,22tayebi2024preserving,mohammadi2026differential}.

\subsubsection{Initialization regimes.}

Recent studies consistently highlight the critical role of initialization in DP-SGD training for medical imaging \cite{arasteh2026roleselfsupervisedpretrainingdifferentially,mohammadi2026differential}. To analyze initialization-dependent geometric responses under DP, we consider three pretrained encoders: (i) supervised ImageNet \cite{deng2009imagenet} initialization as a generic baseline, (ii) self-supervised DINOv3 \cite{simeoni2025dinov3} initialization representing modern foundation models, and (iii) domain-specific initialization pretrained on MIMIC-CXR \cite{johnson2019mimic} (213,921 frontal images), the largest publicly available CXR dataset to date, using identical preprocessing and label space as the downstream task. The architecture and all non-privacy hyperparameters (optimizer, learning rate schedule, batch size, epochs) are fixed across privacy levels and initializations.

\begin{figure}[t]
\centering
\includegraphics[width=0.87\textwidth]{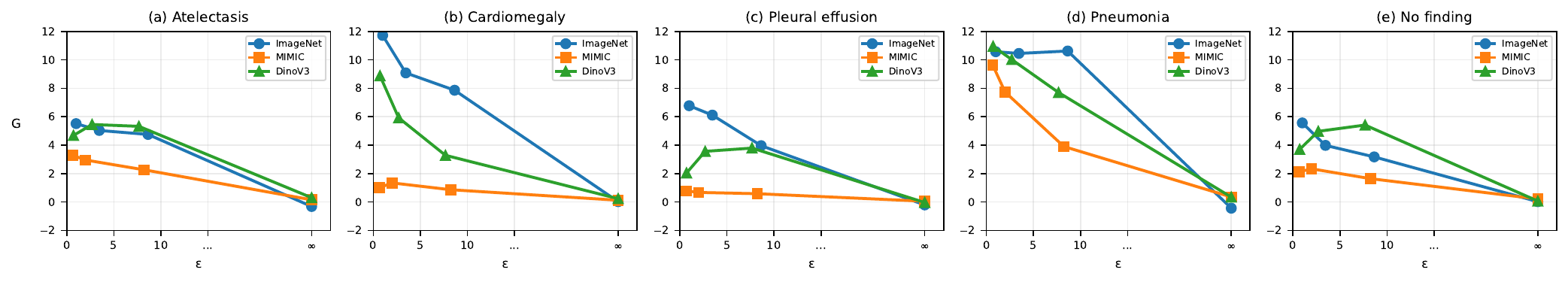}
\caption{Per-label utilization gaps $G(\varepsilon)$ for different $\varepsilon$ on the PadChest dataset.}
\label{fig:perlabel}
\end{figure}

\subsubsection{Statistical estimation.}

Uncertainty is estimated via nonparametric bootstrap over test samples ($B=1000$) \cite{mooney1993bootstrapping}. Within each initialization branch, we compute rank correlations between $\mathrm{AUROC}_{\mathrm{end2end}}$ and geometric statistics across privacy budgets. For each configuration this yields $(\mathrm{AUROC}_{\mathrm{end2end}}$, $\mathrm{AUROC}_{\mathrm{probe}}$, $\Delta(\varepsilon)$, $d_{\mathrm{eff}}(\varepsilon))$, forming the basis of the representation-level analysis. All reported classification metrics are expressed in percent.


\section{Results}

\subsubsection{DP-RGMI decomposes privacy degradation into separability and utilization.}

Table~\ref{tab:macro_results_full} summarizes the results. As expected, $\mathrm{AUROC}_{\mathrm{end2end}}$ decreases under privacy across all initializations (ImageNet: $88.8\rightarrow76.6\rightarrow74.5$; DINOv3: $89.5\rightarrow77.4\rightarrow75.6$; MIMIC: $90.0\rightarrow85.8\rightarrow83.9$ as $\varepsilon$ decreases). However, DP-RGMI asks \emph{where} the degradation arises.

Under non-DP, $G(\infty)\approx0$ for all initializations, indicating that joint training largely realizes the linearly recoverable structure in $\phi_\infty$. Under DP, probe AUROC remains consistently higher than $\mathrm{AUROC}_{\mathrm{end2end}}$, yielding large gaps at strong privacy: $G=8.0$ (ImageNet, $\varepsilon=1.0$), $3.4$ (MIMIC, $\varepsilon=0.7$), and $6.1$ (DINOv3, $\varepsilon=0.7$). This implies that DP can preserve substantial linear separability in $\phi_\varepsilon$ while impairing its utilization during joint DP training.

\begin{table}[t]
\centering
\caption{Overall results on the PadChest dataset, computed by bootstrap on the test set (1000 resamples), reported as mean $\pm$ standard deviation.}
\label{tab:macro_results_full}
\setlength{\tabcolsep}{4.9pt}
\scriptsize
\begin{tabular}{llccccc}
\hline
Initialization & $\varepsilon$ & $\mathrm{AUROC}_{\mathrm{end2end}}$ & $\mathrm{AUROC}_{\mathrm{probe}}$ & $G$ & $\Delta$ & $d_{\mathrm{eff}}$ \\
\hline
\multirow{4}{*}{ImageNet}
& $\infty$ & 88.8 $\pm$ 0.2 & 88.6 $\pm$ 0.2 & -0.2 $\pm$ 0.3 & 1.9 $\pm$ 0.0 & 7.8 $\pm$ 0.1 \\
& 8.6 & 76.6 $\pm$ 0.3 & 82.7 $\pm$ 0.2 & 6.1 $\pm$ 0.4 & 1.1 $\pm$ 0.0 & 3.4 $\pm$ 0.0 \\
& 3.5 & 74.9 $\pm$ 0.3 & 81.9 $\pm$ 0.3 & 6.9 $\pm$ 0.4 & 1.0 $\pm$ 0.0 & 4.7 $\pm$ 0.0 \\
& 1.0 & 74.5 $\pm$ 0.3 & 82.5 $\pm$ 0.3 & 8.0 $\pm$ 0.4 & 1.1 $\pm$ 0.0 & 9.2 $\pm$ 0.1 \\
\hline
\multirow{4}{*}{MIMIC}
& $\infty$ & 90.0 $\pm$ 0.2 & 90.2 $\pm$ 0.2 & 0.2 $\pm$ 0.3 & 0.1 $\pm$ 0.0 & 3.3 $\pm$ 0.0 \\
& 8.2 & 85.8 $\pm$ 0.2 & 87.6 $\pm$ 0.2 & 1.9 $\pm$ 0.3 & 1.4 $\pm$ 0.0 & 4.4 $\pm$ 0.0 \\
& 2.0 & 84.4 $\pm$ 0.3 & 87.4 $\pm$ 0.2 & 3.0 $\pm$ 0.3 & 1.4 $\pm$ 0.0 & 5.4 $\pm$ 0.0 \\
& 0.7 & 83.9 $\pm$ 0.2 & 87.2 $\pm$ 0.2 & 3.4 $\pm$ 0.3 & 1.3 $\pm$ 0.0 & 5.5 $\pm$ 0.0 \\
\hline
\multirow{4}{*}{DINOv3}
& $\infty$ & 89.5 $\pm$ 0.2 & 89.7 $\pm$ 0.2 & 0.2 $\pm$ 0.3 & 0.7 $\pm$ 0.0 & 2.8 $\pm$ 0.0 \\
& 7.7 & 77.4 $\pm$ 0.3 & 82.5 $\pm$ 0.3 & 5.1 $\pm$ 0.4 & 1.9 $\pm$ 0.0 & 5.1 $\pm$ 0.0 \\
& 2.7 & 76.0 $\pm$ 0.3 & 81.9 $\pm$ 0.3 & 6.0 $\pm$ 0.4 & 1.9 $\pm$ 0.0 & 4.6 $\pm$ 0.0 \\
& 0.7 & 75.6 $\pm$ 0.3 & 81.6 $\pm$ 0.3 & 6.1 $\pm$ 0.4 & 1.6 $\pm$ 0.0 & 3.9 $\pm$ 0.0 \\
\hline
\end{tabular}
\end{table}


To test whether the utilization gap reflects a coherent failure mode rather than random degradation, Fig. \ref{fig:perlabel} shows per-label utilization gaps. DP shifts AUROC downward across labels while broadly preserving relative difficulty. This supports a global DP-induced transformation rather than selective erasure of a single label-specific axis. In contrast, the utilization gap becomes sharply label-dependent. Under ImageNet at $\varepsilon=1.0$, pneumonia exhibits a $+10.6$ AUROC gap, whereas no finding has a smaller $+5.6$ gap. Under MIMIC at $\varepsilon=0.7$, gaps are smaller for most labels overall, although pneumonia still shows a relatively large gap ($+9.6$). Under DINOv3 at $\varepsilon=0.7$, pneumonia again shows a large gap ($+11.0$), indicating that utilization failure is not specific to supervised or self-supervised pretraining, but depends on how DP-constrained optimization interacts with initialization and label geometry.


\subsubsection{Geometry under DP.}

DP-RGMI attributes the remaining variation in performance to changes in representation geometry. Table~\ref{tab:macro_results_full} shows that DP induces measurable displacement $\Delta(\varepsilon)$ and reshaping of spectral structure through $d_{\mathrm{eff}}(\varepsilon)$, with patterns that depend on initialization.

\emph{Displacement.}
Under DP, all initializations move away from their pretrained prior, but to different extents. DINOv3 exhibits the largest drift (e.g., $\Delta=1.9$ at $\varepsilon=7.7$), ImageNet shows moderate but consistent displacement ($\Delta\approx1.0$--$1.1$), and MIMIC transitions from near-zero movement without privacy ($\Delta=0.1$) to substantial displacement under DP ($\Delta\approx1.3$--$1.4$). Importantly, displacement magnitude does not map monotonically to utility. For example, configurations with similar AUROC can correspond to different $\Delta$ values, indicating that geometric departure from initialization alone does not determine task performance.

\begin{table}[t]
\centering
\caption{Spearman rank correlation $\rho$ with $\mathrm{AUROC}_{\mathrm{end2end}}$ for DP models ($\epsilon<10$). Across inits: $n=3\times3$ ($\epsilon$, datasets); across datasets: $n=3\times3$ ($\epsilon$, inits).}
\label{tab:corr_summary}
\setlength{\tabcolsep}{3.2pt}
\scriptsize
\begin{tabular}{lcccc}
\hline
Setting & $n$
& $\rho(\mathrm{AUROC}_{\mathrm{end2end}},G)$
& $\rho(\mathrm{AUROC}_{\mathrm{end2end}},\Delta)$
& $\rho(\mathrm{AUROC}_{\mathrm{end2end}},d_{\mathrm{eff}})$ \\
\hline
ImageNet init & 9 & -0.78 & -0.43 & -0.33 \\
MIMIC init    & 9 & -0.31 & +0.81 & +0.49 \\
DINOv3 init   & 9 & +0.55 & +0.15 & +0.52 \\
\hline
PadChest dataset       & 9 & -0.95 & +0.32 & -0.07 \\
CheXpert dataset       & 9 & -0.86 & -0.39 & -0.02 \\
ChestX-ray14 dataset   & 9 & -0.98 & -0.23 & -0.43 \\
\hline
\textbf{Overall} & \textbf{27}
& \textbf{-0.61}
& \textbf{-0.12}
& \textbf{-0.07} \\
\hline
\end{tabular}
\end{table}

\emph{Spectral reshaping.}
Changes in $d_{\mathrm{eff}}(\varepsilon)$ are non-monotonic and initialization-dependent. Under ImageNet, $d_{\mathrm{eff}}$ decreases at moderate privacy ($3.4$ at $\varepsilon=8.6$) but increases at stronger privacy ($9.2$ at $\varepsilon=1.0$). In contrast, DINOv3 trends toward lower effective dimension as privacy strengthens ($5.1 \rightarrow 3.9$), while MIMIC exhibits a gradual increase under DP. These heterogeneous trajectories argue against a uniform representation collapse. Instead, DP induces structured spectral transformations whose direction depends on the pretrained prior. Geometry therefore provides context for how privacy reshapes embeddings, while the utilization gap identifies where performance is lost.


\subsubsection{Cross-dataset generalization and correlation structure.}

We next examine the generalization beyond PadChest. The full protocol is repeated on CheXpert \cite{irvin2019chexpert} (total of 157,676 frontal images)  and ChestX-ray14 \cite{wang2017chestx} (total of 112,120 frontal images) datasets under identical data partitioning ($15-23\%$ patient-stratified held-out test sets), $\delta$, initialization regimes, training settings, and statistical estimation, with comparable privacy budgets. Datasets are evaluated as multi-label classification for the same 5 labels, with macro-averaged analysis.

\begin{figure}[t]
\centering
\includegraphics[width=0.8\textwidth]{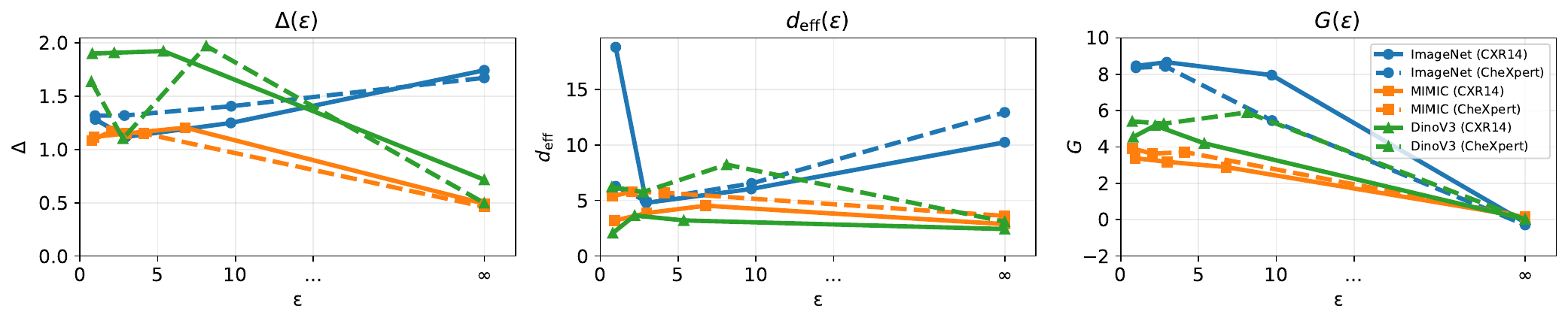}
\caption{Generalization results on CheXpert and ChestX-ray14 datasets.}
\label{fig:generalization}
\end{figure}

Fig.~\ref{fig:generalization} shows the trajectories of $G(\varepsilon)$, $\Delta(\varepsilon)$, and $d_{\mathrm{eff}}(\varepsilon)$ for both generalization datasets. Despite differences in dataset size and baseline AUROC, a consistent pattern emerges: under stronger privacy, probe AUROC remains higher than $\mathrm{AUROC}_{\mathrm{end2end}}$ across initializations, yielding a positive $G(\varepsilon)$. In contrast, $\Delta(\varepsilon)$ and $d_{\mathrm{eff}}(\varepsilon)$ follow dataset- and initialization-specific trajectories rather than exhibiting uniform degradation.
To quantify these relationships, we compute Spearman rank correlations between $\mathrm{AUROC}_{\mathrm{end2end}}$ and DP-RGMI quantities for DP models (Table~\ref{tab:corr_summary}). Across initializations, the association between AUROC and $G$ is negative for ImageNet ($\rho=-0.78$) but becomes weak or reverses sign for MIMIC ($\rho=-0.31$) and DINOv3 ($\rho=+0.55$), indicating that the monotonic relationship is initialization-dependent within the DP regime. In contrast, geometric associations can be substantial for some priors (e.g., MIMIC: $\rho=+0.81$ with $\Delta$), suggesting that geometry captures prior-conditioned variation not explained by $G$ alone.
Across datasets, AUROC remains negatively associated with $G$ (PadChest: $\rho=-0.95$, CheXpert: $\rho=-0.86$, ChestX-ray14: $\rho=-0.98$), while correlations with $\Delta$ and $d_{\mathrm{eff}}$ remain dataset-specific. Overall, the association between AUROC and $G$ is moderate ($\rho=-0.61$), whereas correlations with $\Delta$ ($\rho=-0.12$) and $d_{\mathrm{eff}}$ ($\rho=-0.07$) are weak. Note that the association with $G$ partly reflects its definition relative to AUROC and is interpreted descriptively rather than causally.

\section{Discussion and conclusion}

We reframed DP evaluation in CXR classification as a representation-level diagnostic problem. 
Instead of relying solely on end-to-end performance, DP-RGMI separates encoder geometry from downstream utilization. Strong privacy is consistently associated with a utilization gap, while the strength of this relationship can vary by initialization and geometry can explain additional prior- and dataset-conditioned variation.

This separation supports concrete deployment decisions. If two privacy budgets yield similar AUROC but one exhibits a larger $G$, DP-RGMI indicates that recoverable signal persists, pointing to optimization-side remedies (freezing the encoder, retraining only the head, or adjusting clipping for head parameters) as candidate interventions to be tested rather than relaxing privacy. Our linear-probe measurement already realizes a controlled instance of this: $U_{\mathrm{probe}}>U_{\mathrm{end2end}}$ under DP (Table~\ref{tab:macro_results_full}) shows the recoverable structure is exploitable once the head is trained on a frozen private encoder, and quantifying the privacy cost of such head-only re-optimization is direct future work. If $\Delta$ is large while probe performance remains stable, representation has moved substantially from its pretrained prior, which may affect transfer or reuse across institutions even when classification performance appears acceptable. Conversely, marked reductions in $d_{\mathrm{eff}}$ indicate increased spectral concentration and reduced representational diversity, potentially limiting adaptation to new tasks. In such cases, revisiting pretraining or privacy strength may be more appropriate than head-level adjustments.

In this study, all experiments use multi-label chest X-ray classification with a single ConvNeXt-Small backbone. This choice is constrained rather than free: batch normalization is incompatible with the per-sample gradients of DP-SGD, and transformers exhibit unstable DP optimization \cite{mohammadi2026differential}, while cross-architecture consistency of DP trade-offs has been reported separately \cite{mohammadi2026differential,13TayebiDomainTransfer,22tayebi2024preserving,67ziller2024reconciling}. The framework is model-agnostic by construction; its behavior in other tasks such as segmentation remains to be validated, and we expect similar geometry-utilization interactions wherever representations are reused or fine-tuned.

Overall, DP-RGMI provides a reproducible framework for diagnosing privacy-induced failure modes and guiding principled privacy model selection in cases with cross-institutional reuse, transfer learning, or frozen-feature deployment.


\begin{credits}
\subsubsection{\ackname}
STA is supported by the Excellence Strategy of the German Federal Government, the L\"ander, and RWTH ERS (START\_526-26). SN is supported by the DFG (701010997, 517243167). DT is supported by the German Ministry of Research, Technology and Space (TRANSFORM LIVER - 031L0312C, DECIPHER-M - 01KD2420B), DFG (515639690), and the European Union (Horizon Europe, ODELIA - GA 101057091, ERC Starting Grant SAGMA -- GA 101222556).

\subsubsection{\discintname}
DT received honoraria for lectures by Bayer, GE, Roche, AstraZeneca, and Philips and holds shares in StratifAI GmbH and Synagen GmbH, Germany. The other authors declare no competing interests relevant to this work.
\end{credits}

\bibliographystyle{splncs04}
\bibliography{paper}

@article{mohammadi2026differential,
  title={Differential privacy for medical deep learning: methods, tradeoffs, and deployment implications},
  author={Mohammadi, Marziyeh and Vejdanihemmat, Mohsen and Lotfinia, Mahshad and Rusu, Mirabela and Truhn, Daniel and Maier, Andreas and Tayebi Arasteh, Soroosh},
  journal={npj Digit. Med.},
  year={2026},
  volume={9},
  pages={93},
  publisher={Nature Publishing Group UK London}
}

@misc{dwork2014algorithmic,
  title={The algorithmic foundations of differential privacy. foundations and trends{\textregistered} in theoretical computer science 9 (3-4), 211--407 (2014)},
  author={Dwork, Cynthia and Roth, Aaron}
}

@article{usynin2021adversarial,
  title={Adversarial interference and its mitigations in privacy-preserving collaborative machine learning},
  author={Usynin, Dmitrii and Ziller, Alexander and Makowski, Marcus and Braren, Rickmer and Rueckert, Daniel and Glocker, Ben and Kaissis, Georgios and Passerat-Palmbach, Jonathan},
  journal={Nat Mach Intell},
  volume={3},
  number={9},
  pages={749--758},
  year={2021},
  publisher={Nature Publishing Group UK London}
}

@article{vaswani2017attention,
  title={Attention is all you need},
  author={Vaswani, A},
  journal={NeurIPS 2017},
 volume = {30}
 }

@inproceedings{DPSGD,
  title={Deep learning with differential privacy},
  author={Abadi, Martin and Chu, Andy and Goodfellow, Ian and McMahan, H Brendan and Mironov, Ilya and Talwar, Kunal and Zhang, Li},
  booktitle={SIGSAC 2016},
  pages={308--318},
}

@article{13TayebiDomainTransfer,
  title={Securing collaborative medical AI by using differential privacy: Domain transfer for classification of chest radiographs},
  author={Tayebi Arasteh, Soroosh and Lotfinia, Mahshad and Nolte, Teresa and S{\"a}hn, Marwin-Jonathan and Isfort, Peter and Kuhl, Christiane and Nebelung, Sven and Kaissis, Georgios and Truhn, Daniel},
  journal={Radiology: Artificial Intelligence},
  volume={6},
  number={1},
  pages={e230212},
  year={2023},
  publisher={Radiological Society of North America}
}

@article{22tayebi2024preserving,
  title={Preserving fairness and diagnostic accuracy in private large-scale AI models for medical imaging},
  author={Tayebi Arasteh, Soroosh and Ziller, Alexander and Kuhl, Christiane and Makowski, Marcus and Nebelung, Sven and Braren, Rickmer and Rueckert, Daniel and Truhn, Daniel and Kaissis, Georgios},
  journal={Commun Med},
  volume={4},
  number={1},
  pages={46},
  year={2024},
  publisher={Nature Publishing Group UK London}
}

@article{66kaissis2021end,
  title={End-to-end privacy preserving deep learning on multi-institutional medical imaging},
  author={Kaissis, Georgios and Ziller, Alexander and Passerat-Palmbach, Jonathan and Ryffel, Th{\'e}o and Usynin, Dmitrii and Trask, Andrew and Lima Jr, Ion{\'e}sio and Mancuso, Jason and Jungmann, Friederike and Steinborn, Marc-Matthias and others},
  journal={Nat Mach Intell},
  volume={3},
  number={6},
  pages={473--484},
  year={2021},
  publisher={Nature Publishing Group UK London}
}

@article{67ziller2024reconciling,
  title={Reconciling privacy and accuracy in AI for medical imaging},
  author={Ziller, Alexander and Mueller, Tamara T and Stieger, Simon and Feiner, Leonhard F and Brandt, Johannes and Braren, Rickmer and Rueckert, Daniel and Kaissis, Georgios},
  journal={Nat Mach Intell},
  volume={6},
  number={7},
  pages={764--774},
  year={2024},
  publisher={Nature Publishing Group UK London}
}

@article{72arasteh2024differentialprivacyenablesfair,
  title={Differential privacy enables fair and accurate AI-based analysis of speech disorders while protecting patient data},
  author={Tayebi Arasteh, Soroosh and Lotfinia, Mahshad and Perez-Toro, Paula Andrea and others},
  journal={npj Artif. Intell.},
  volume={1},
  pages={37},
  year={2025},
  publisher={Nature Publishing Group UK London}
}

@article{bustos2020padchest,
  title={Padchest: A large chest x-ray image dataset with multi-label annotated reports},
  author={Bustos, Aurelia and Pertusa, Antonio and Salinas, Jose-Maria and De La Iglesia-Vaya, Maria},
  journal={Medical image analysis},
  volume={66},
  pages={101797},
  year={2020},
  publisher={Elsevier}
}

@article{johnson2019mimic,
  title={MIMIC-CXR, a de-identified publicly available database of chest radiographs with free-text reports},
  author={Johnson, Alistair EW and Pollard, Tom J and Berkowitz, Seth J and Greenbaum, Nathaniel R and Lungren, Matthew P and Deng, Chih-ying and Mark, Roger G and Horng, Steven},
  journal={Sci Data},
  volume={6},
  pages={317},
  year={2019},
  publisher={Nature Publishing Group UK London}
}

@inproceedings{deng2009imagenet,
  title={Imagenet: A large-scale hierarchical image database},
  author={Deng, Jia and Dong, Wei and Socher, Richard and Li, Li-Jia and Li, Kai and Fei-Fei, Li},
  booktitle={CVPR 2009},
  pages={248--255},
}

@article{simeoni2025dinov3,
  title={Dinov3},
  author={Sim{\'e}oni, Oriane and Vo, Huy V and Seitzer, Maximilian and Baldassarre, Federico and Oquab, Maxime and Jose, Cijo and Khalidov, Vasil and Szafraniec, Marc and Yi, Seungeun and Ramamonjisoa, Micha{\"e}l and others},
  journal={arXiv preprint arXiv:2508.10104},
  year={2025}
}

@InProceedings{Liu_2022_CVPR,
    author    = {Liu, Zhuang and Mao, Hanzi and Wu, Chao-Yuan and Feichtenhofer, Christoph and Darrell, Trevor and Xie, Saining},
    title     = {A ConvNet for the 2020s},
    booktitle = {CVPR 2022},
    pages     = {11976-11986}
}

@article{arasteh2026roleselfsupervisedpretrainingdifferentially,
  title={The pretraining domain outweighs the training objective in setting the privacy-utility trade-off of differentially private medical image analysis},
  author={Soroosh Tayebi Arasteh and Mina Farajiamiri and Mahshad Lotfinia and others},
  journal={arXiv preprint arXiv:2601.19618},
  year={2026}
}

@book{mooney1993bootstrapping,
  author = {Mooney, Christopher Z. and Duval, Robert D.},
  isbn = {9780803953819   080395381X},
  series = {Quantitative Applications in the Social Sciences},
  title = {Bootstrapping: A Nonparametric Approach to Statistical Inference},
  year = 1993
}

@inproceedings{NEURIPS2019_cfcce062,
 author = {Ansuini, Alessio and Laio, Alessandro and Macke, Jakob H and Zoccolan, Davide},
 booktitle = {NeurIPS 2019},
 title = {Intrinsic dimension of data representations in deep neural networks},
 volume = {32},
}

@inproceedings{10555535249383525087,
author = {Chen, Ting and Kornblith, Simon and Norouzi, Mohammad and Hinton, Geoffrey},
title = {A simple framework for contrastive learning of visual representations},
booktitle = {ICML 2020},
articleno = {149},
}

@inproceedings{wang2017chestx,
  title={Chestx-ray8: Hospital-scale chest x-ray database and benchmarks on weakly-supervised classification and localization of common thorax diseases},
  author={Wang, Xiaosong and Peng, Yifan and Lu, Le and Lu, Zhiyong and Bagheri, Mohammadhadi and Summers, Ronald M},
  booktitle={CVPR 2017},
  pages={2097--2106},
}

@inproceedings{irvin2019chexpert,
  title={Chexpert: A large chest radiograph dataset with uncertainty labels and expert comparison},
  author={Irvin, Jeremy and Rajpurkar, Pranav and others},
  booktitle={Proceedings of the AAAI conference on artificial intelligence},
  volume={33},
  number={01},
  pages={590--597},
  year={2019}
}

@inproceedings{NEURIPS2021_6097d8f3,
 author = {Gopi, Sivakanth and Lee, Yin Tat and Wutschitz, Lukas},
 booktitle = {NeurIPS},
 pages = {11631--11642},
 title = {Numerical Composition of Differential Privacy},
 volume = {34},
 year = {2021}
}

\end{document}